\documentclass[journal]{IEEEtran}

\usepackage[utf8]{inputenc}
\usepackage{color}
\usepackage{xcolor}
\usepackage{array}
\usepackage{verbatim}
\usepackage{float}
\usepackage{amsmath}
\usepackage{amsthm}
\usepackage{amssymb}
\usepackage{graphicx}
\usepackage{longtable}
\usepackage{multirow}
\usepackage{booktabs}
\usepackage[unicode=true,
bookmarks=false,
breaklinks=false,pdfborder={0 0 1},colorlinks=false]
{hyperref}
\hypersetup{
	colorlinks,bookmarksopen,bookmarksnumbered,citecolor=blue,urlcolor=blue}
\usepackage{cite}

\usepackage{lipsum}
\usepackage{mathtools}
\usepackage{cuted}
\providecommand{\tabularnewline}{\\}
\usepackage{algorithmic}
\usepackage{longtable}

\floatstyle{ruled}
\newfloat{algorithm}{tbp}{loa}
\providecommand{\algorithmname}{Algorithm}
\floatname{algorithm}{\protect\algorithmname}

\makeatletter
\let\oldforeign@language\foreign@language
\DeclareRobustCommand{\foreign@language}[1]{%
	\lowercase{\oldforeign@language{#1}}}

\let\oldforeign@language\foreign@language
\DeclareRobustCommand{\foreign@language}[1]{%
	\lowercase{\oldforeign@language{#1}}}

\ifCLASSINFOpdf
\else
\fi

\@ifundefined{showcaptionsetup}{}{%
	\PassOptionsToPackage{caption=false}{subfig}}
\usepackage{subfig}

\usepackage{balance}

\ifCLASSINFOpdf
\else
\fi

\ifCLASSINFOpdf
\else
\fi

\makeatother
\begin{document}
	\bstctlcite{IEEEexample:BSTcontrol}

\title{Unscented Particle Filter for Visual-inertial Navigation using IMU and Landmark Measurements}

\author{Khashayar Ghanizadegan and Hashim A. Hashim
\thanks{This work was supported in part by National Sciences and Engineering Research Council of Canada (NSERC), under the grants RGPIN-2022-04937 and DGECR-2022-00103.} 
	\thanks{K. Ghanizadegan and H. A. Hashim are with the Department of Mechanical
		and Aerospace Engineering, Carleton University, Ottawa, Ontario, K1S-5B6,
		Canada (e-mail: hhashim@carleton.ca).}
}



\maketitle
\begin{abstract}
This paper introduces a geometric Quaternion-based Unscented Particle
Filter for Visual-Inertial Navigation (QUPF-VIN) specifically designed
for a vehicle operating with six degrees of freedom (6 DoF). The proposed
QUPF-VIN technique is quaternion-based capturing the inherently nonlinear
nature of true navigation kinematics. The filter fuses data from a
low-cost inertial measurement unit (IMU) and landmark observations
obtained via a vision sensor. The QUPF-VIN is implemented in discrete
form to ensure seamless integration with onboard inertial sensing
systems. Designed for robustness in GPS-denied environments, the proposed
method has been validated through experiments with real-world dataset
involving an unmanned aerial vehicle (UAV) equipped with a 6-axis
IMU and a stereo camera, operating with 6 DoF. The numerical results
demonstrate that the QUPF-VIN provides superior tracking accuracy
compared to ground truth data. Additionally, a comparative analysis
with a standard Kalman filter-based navigation technique further highlights
the enhanced performance of the QUPF-VIN.
\end{abstract}

\section{Introduction}\label{sec1}

\IEEEPARstart{A}{ccurate} navigation algorithms are crucial for autonomous ground and
aerial vehicles, particularly in both indoor and outdoor applications
where Global Positioning System (GPS) signals may be unavailable or
unreliable. These techniques are essential for tasks such as warehouse
management, surveillance, road and pipeline inspection, package delivery,
and household robotics \cite{hashim2025advances,koch2020relative,hashim2021geometricNAV}.
In such cases, GPS-independent navigation solutions become critical.
In GPS-denied environments, autonomous vehicles rely on robust algorithms
capable of providing reliable estimates using cost-effective inertial
measurement units (IMUs). Navigation can be achieved with low-cost
onboard sensors, such as 6-axis IMUs and either stereo or monocular
cameras \cite{hashim2025advances,hashim2021geometricNAV}. A 6-axis
IMU, comprising a gyroscope and accelerometer, provides measurements
of a vehicle's angular velocity and linear acceleration. Dead-reckoning,
which has been extensively studied, is commonly used to estimate a
vehicle’s navigation state (attitude (orientation), position, and
linear velocity), while operating with six degrees of freedom (6 DoF)
\cite{hashim2025advances,hashim2021geometricNAV}. This technique
relies solely on IMU data, using numerical integration based on the
vehicle’s initial state. However, dead-reckoning is prone to error
accumulation, making it unsuitable for long-distance navigation \cite{hashim2021geometricNAV}.
Attitude-only estimation, on the other hand, can be reliably achieved
through IMU measurements using robust attitude filters, including
deterministic filters \cite{hashim2021PosePPF} and stochastic approaches
\cite{hashim2021gps,hashim2021geometricNAV,hashim2019Ito}.

Pose estimation (attitude and position) of a vehicle navigating in
three-dimensional (3D) space can be achieved through sensor fusion,
such as the integration of landmark measurements from a vision system
and data from an IMU. Commonly used filters for pose estimation include
Kalman filters, the Extended Kalman Filter (EKF), and nonlinear filters
\cite{ha2022Pose,vasconcelos2010discrete}. However, these algorithms
typically require knowledge of the vehicle's linear velocity, which
poses a significant challenge in GPS-denied environments. In practice,
uncertain attitude and position can be reconstructed through the fusion
of vision data and IMU measurements \cite{fornasier2022equivariant}.
Nevertheless, deriving or optimizing linear velocity from reconstructed
uncertain attitude and position data proves impractical and unreliable.
As a result, there is a growing demand for robust navigation techniques
that can handle measurement uncertainties, provide accurate attitude
and position estimates, and observe the vehicle's linear velocity,
which is often considered a hidden state \cite{hashim2021geometricNAV}.
True 6 DoF navigation kinematics are highly nonlinear \cite{hashim2021geometricNAV},
making linear filters inadequate for accurate navigation estimation
\cite{hashim2021geometricNAV,Hashim2023NonlinearFusion}. Kalman filters
(KF) have been applied to vision-based navigation \cite{ali2009consistent},
with modifications such as the EKF \cite{ali2009consistent} introduced
to account for system nonlinearity. Further enhancements, like the
Multiplicative EKF (MEKF) \cite{koch2020relative} and the Multi-State
Constraint Kalman Filter (MSCKF) \cite{sun2018robust}, were developed
to improve accuracy and address consistency issues. Additionally,
Unscented Kalman Filters (UKF) \cite{cantelobre2020real} have been
proposed to better address the nonlinearity of kinematic models. However,
the main limitation of KF is its disregard for navigation nonlinearities,
while EKF linearizes the system around a nominal point. MEKF, MSCKF,
and UKF rely on parametric statistical models that fail to capture
the full complexity of arbitrary distributions in nonlinear navigation
systems. The Particle Filter (PF) \cite{hong2021particle}, which
can capture arbitrary distributions, has been applied to navigation
problems but struggles numerically when dealing with relatively accurate
sensors. These sensors generate narrow distributions, causing particles
to receive near-zero probabilities if they deviate slightly from the
distribution's peak, thus hindering effective guidance. The Unscented
PF (UPF) \cite{van2000unscented} addresses this limitation by using
UKF to propagate particles and estimate their mean and covariance,
resulting in better alignment with the posterior distribution. This
approach overcomes the shortcomings of PF when dealing with narrow
distributions \cite{van2000unscented}.

\paragraph*{Contribution}

This paper presents a geometric Quaternion-based Unscented Particle
Filter for Visual-Inertial Navigation (QUPF-VIN) that relies on sensor
data from a 6-axis IMU and landmark measurements. The proposed filter
captures the nonlinear nature of navigation kinematics and is designed
and implemented in a geometric discrete form at a low sampling rate.
The effectiveness of QUPF-VIN is validated using a real-world quadrotor
dataset, which includes low-cost IMU data and stereo images, and is
compared against ground-truth data. The proposed navigation algorithm
is suitable for implementation in both Unmanned Aerial Vehicles (UAVs)
and ground vehicles.

\paragraph*{Structure}

The remainder of the paper is organized as follows: Section \ref{sec:Preliminaries-and-Math}
discusses important math notation and preliminaries. Section \ref{sec:Problem-Formulation}
presents the true navigation kinematics and inertial measurement data
(IMU and landmarks). Section \ref{sec:Filter} introduces the QUPF-VIN,
and steps of implementation. Section \ref{sec:Results} illustrates
the output performance of the proposed QUPF-VIN using a dataset obtained
from real quadrotor flight and compares QUPF-VIN performance to a
baseline filter. Finally, Section \ref{sec:Conclusion} provides concluding
remarks.

\section{Preliminary and Math Notation\label{sec:Preliminaries-and-Math}}

\begin{table}[t]
	\centering{}\caption{\label{tab:Table-of-Notations2}Nomenclature}
	\begin{tabular}{>{\raggedright}p{2cm}l>{\raggedright}p{5.6cm}}
		\toprule 
		\addlinespace
		$\left\{ \mathcal{B}\right\} $ / $\left\{ \mathcal{W}\right\} $  & :  & Fixed body-frame / fixed world-frame\tabularnewline
		\addlinespace
		$\mathbb{SO}\left(3\right)$  & :  & Special Orthogonal Group of order 3\tabularnewline
		\addlinespace
		$\mathbb{S}^{3}$  & :  & Three-unit-sphere\tabularnewline
		\addlinespace
		$\mathbb{R}^{n_{1}\times n_{2}}$  & :  & $n_{1}$-by-$n_{2}$ real space\tabularnewline
		\addlinespace
		$q_{k},\hat{q}_{k}$  & :  & True and estimated quaternion at step $k$\tabularnewline
		\addlinespace
		$p_{k},\hat{p}_{k}$  & :  & True and estimated position at step $k$\tabularnewline
		\addlinespace
		$v_{k},\hat{v}_{k}$  & :  & True and estimated linear velocity at step $k$\tabularnewline
		\addlinespace
		$r_{e,k}$, $p_{e,k}$, $v_{e,k}$  & :  & Attitude, position, and velocity estimation error\tabularnewline
		\addlinespace
		$a_{k},a_{m,k}$  & :  & True and measured acceleration at step $k$\tabularnewline
		\addlinespace
		$\omega_{k},\omega_{m,k}$  & :  & True and measured angular velocity at step $k$\tabularnewline
		\addlinespace
		$n_{\omega,k},n_{a,k}$  & :  & Angular velocity and acceleration measurements noise\tabularnewline
		\addlinespace
		$b_{\omega,k},b_{a,k}$  & :  & Angular velocity and acceleration measurements bias\tabularnewline
		\addlinespace
		$C_{\times}$  & :  & Covariance matrix of ${\times}$.\tabularnewline
		\addlinespace
		$f_{b},f_{w}$  & :  & landmark points coordinates in $\left\{ \mathcal{B}\right\} $ and
		$\left\{ \mathcal{W}\right\} $.\tabularnewline
		\addlinespace
		$x_{k}$, $u_{k}$  & :  & The state, and input vectors at the $k$th time step\tabularnewline
		\addlinespace
		$z_{k}$  & :  & True measurement\tabularnewline
		\addlinespace
		$\{\mathcal{X}_{k|l}^{(i)}\}$, $\{\mathcal{X}_{k|l}^{(i)a}\}$, $\{\mathcal{Z}_{k|l}\}$  & :  & Sigma points of state, augmented state, and measurements\tabularnewline
		\addlinespace
		$\{\mathcal{X}_{k}^{(i)}\}$  & :  & Particles at step $k$\tabularnewline
		\bottomrule
	\end{tabular}
\end{table}

The set of $n_{1}$-by-$n_{2}$ real number matrices are described
by $\mathbb{R}^{n_{1}\times n_{2}}$. For a vector $x\in\mathbb{R}^{n_{x}}$,
the Euclidean norm is denoted by $\|x\|=\sqrt{x^{\top}x}\in\mathbb{R}$.
$\mathbf{I}_{n}\in\mathbb{R}^{n\times n}$ denotes an identity matrix.
$\left\{ \mathcal{W}\right\} $ describes the world-frame fixed to
the Earth while $\left\{ \mathcal{B}\right\} $ refers to the body-frame
fixed to a moving vehicle. Table \ref{tab:Table-of-Notations2} lists
a set of important symbols used subsequently. $[x]_{\times}$ denotes
skew-symmetric of $x\in\mathbb{R}^{3}$ such that
\begin{equation}
	[x]_{\times}=\left[\begin{array}{ccc}
		0 & -x_{3} & x_{2}\\
		x_{3} & 0 & -x_{1}\\
		-x_{2} & x_{1} & 0
	\end{array}\right]\in\mathfrak{so}(3),\hspace{1em}x=\left[\begin{array}{c}
		x_{1}\\
		x_{2}\\
		x_{3}
	\end{array}\right]
\end{equation}
${\rm vex}(\cdot)$ describes the inverse mapping where ${\rm vex}([x]_{\times})=x\in\mathbb{R}^{3}$.
For $D\in\mathbb{R}^{3\times3}$, the anti-symmetric projection operator
is given by:
\begin{equation}
	\mathcal{P}_{a}(D)=\frac{1}{2}(D-D^{\top})\in\mathfrak{so}(3)\label{eq:pa}
\end{equation}
Orientation of a vehicle is denoted by $R\in\mathbb{SO}(3)$ where
$\mathbb{SO}(3)$ refers to the Special Orthogonal Group such that
\cite{hashim2019special}:
\begin{equation}
	\mathbb{SO}(3):=\left\{ \left.R\in\mathbb{R}^{3\times3}\right|det(R)=+1,R^{\top}R=\mathbf{I}_{3}\right\} 
\end{equation}
Unit-quaternion $q=[q_{w},q_{x},q_{y},q_{z}]^{\top}=[q_{w},q_{v}^{\top}]^{\top}\in\mathbb{S}^{3}$
where $q_{v}\in\mathbb{R}^{3}$ and $q_{w}\in\mathbb{R}$ can be used
to describe the vehicle's orientation where
\begin{equation}
	\mathbb{S}^{3}:=\left\{ q\in\left.\mathbb{R}^{4}\right\Vert |q\|=1\right\} 
\end{equation}
and the vehicle's orientation is given by \cite{hashim2019special}:
\begin{equation}
	R_{q}(q)=(q_{w}^{2}-\|q_{v}\|^{2})I_{3}+2q_{v}q_{v}^{\top}+2q_{w}[q_{v}]_{\times}\in\mathbb{SO}(3)\label{Q2R}
\end{equation}
Define $\otimes$ as quaternion product of two quaternions such that
the quaternion product of $q_{1}=[q_{w1},q_{v1}]^{\top}\in\mathbb{S}^{3}$
and $q_{2}=[q_{w2},q_{v2}]^{\top}\in\mathbb{S}^{3}$ is \cite{hashim2019special}:
\begin{align}
	q_{3} & =q_{1}\otimes q_{2}\nonumber \\
	& =\begin{bmatrix}q_{w1}q_{w2}-q_{v1}^{\top}q_{v2}\\
		q_{w1}q_{v2}+q_{w2}q_{v1}+[q_{v1}]_{\times}q_{v2}
	\end{bmatrix}\in\mathbb{S}^{3}\label{eq:qxq}
\end{align}
and the inverse quaternion of $q=[q_{w},q_{v}^{\top}]^{\top}\in\mathbb{S}^{3}$
is defined by $q^{-1}=[q_{w},-q_{v}^{\top}]^{\top}\in\mathbb{S}^{3}$.
$q_{I}=[1,0,0,0]^{\top}$ describes the quaternion identity such that
$q\otimes q^{-1}=q_{I}$. Angle-axis parameterization describes the
orientation as a rotation (angle) $\alpha\in\mathbb{R}$ around a
unit vector (axis) $v=[v_{1},v_{2},v_{3}]\in\mathbb{S}^{2}$ such
that
\begin{equation}
	r=r_{\alpha,v}(\alpha,v)=\alpha v\in\mathbb{R}^{3}\label{AA2r}
\end{equation}
where the rotation matrix corresponding to the angle-axis parameterization
is defined by
\begin{align}
	R_{r}(r) & =\exp([r]_{\times})\in\mathbb{SO}(3)\nonumber \\
	& =\mathbf{I}_{3}+\sin(\alpha)\left[v\right]_{\times}+\left(1-\cos(\alpha)\right)\left[v\right]_{\times}^{2}\label{r2R}
\end{align}
Note that $\alpha_{R}=\arccos\left(\frac{{\rm Tr}(R)-1}{2}\right)\in\mathbb{R}$
and $v_{R}=\frac{1}{\sin\alpha_{R}}{\rm vex}(\mathcal{P}_{a}(R))\in\mathbb{S}^{2}$
\cite{hashim2019special}. Using the rotation vector $r$ in \eqref{AA2r},
one obtains
\begin{align}
	q_{r}(r) & =q_{R}\left(R_{r}(r)\right)\in\mathbb{S}^{3}\nonumber \\
	& =\left[\cos(\alpha/2),\sin(\alpha/2)v^{\top}\right]^{\top}\in\mathbb{S}^{3}\label{eq:r2q}
\end{align}
with
\begin{equation}
	r_{q}(q)=r_{\alpha,v}(\alpha_{R}(R_{q}(q)),v_{R}(R_{q}(q)))\in\mathbb{R}^{3}\label{eq:q2r}
\end{equation}
The following quaternion subtraction operator is defined:
\begin{equation}
	q_{1}\ominus q_{2}:=r_{q}(q_{1}\otimes q_{2}^{-1})\in\mathbb{R}^{3},\hspace{1em}\forall q_{1},q_{2}\in\mathbb{S}^{3}\label{eq:q-q}
\end{equation}
And the quaternion-rotation vector addition and subtraction operators
are defined as:
\begin{align}
	q\oplus r & :=q_{r}(r)\otimes q\in\mathbb{S}^{3},\hspace{1.4cm}\forall q\in\mathbb{S}^{3},\forall r\in\mathbb{R}^{3}\label{eq:q+r}\\
	q\ominus r & :=q_{r}(r)^{-1}\otimes q\in\mathbb{S}^{3},\hspace{1cm}\forall q\in\mathbb{S}^{3},\forall r\in\mathbb{R}^{3}\label{eq:q-r}
\end{align}
Consider $S=\{s_{i}\}$ to be scaler weights and let us define the
following term:
\[
D=\sum s_{i}q_{i}q_{i}^{\top}\in\mathbb{R}^{4\times4}
\]
The unit eigenvector associated to eigenvalue with the highest magnitude
can be obtained by:
\begin{equation}
	{\rm WM}(Q,W)=\text{EigVector}(D)_{i}\in\mathbb{S}^{3}\label{eq:weighted average}
\end{equation}
where $i=\text{argmax}(|\text{EigValue}(D)_{i}|)\in\mathbb{R}$, ${\rm WM}(D,W)$
denotes weighted mean, and $\text{EigValue}(D)_{i}$ refers to the
$i$th eigenvalue of $D$. The Gaussian probability density function
of $a$ is defined as follows:
\begin{align*}
	\mathbb{P}(a) & =\mathcal{N}(a|\overline{a},P_{a})\\
	& =\frac{\exp\left(-\frac{1}{2}(a-\overline{a})^{\top}P_{a}^{-1}(a-\overline{a})\right)}{\sqrt{(2\pi)^{n}\det(P_{a})}}\in\mathbb{R}
\end{align*}
where $a\in\mathbb{R}^{n}$ obtained through a Gaussian distribution
$a\sim\mathcal{N}(\overline{a},P_{a})$, $\overline{a}\in\mathbb{R}^{n}$
denotes the mean of $a$, and $P_{a}\in\mathbb{R}^{n\times n}$ describes
covariance matrix related to $a$.

\section{Problem Formulation and Sensor Data\label{sec:Problem-Formulation}}

\subsection{Navigation Model}

Consider a vehicle travelling in 3D space where $\omega\in\mathbb{R}^{3}$
denotes its angular velocity and $a\in\mathbb{R}^{3}$ describes its
acceleration measured with $\omega,a\in\{\mathcal{B}\}$. The vehicle's
position and linear velocity are described by $p\in\mathbb{R}^{3}$
and $v\in\mathbb{R}^{3}$, respectively, where $p,v\in\{\mathcal{W}\}$,
whereas the vehicle's orientation is described in view of quaternion
$q\in\mathbb{S}^{3}$ and $q\in\{\mathcal{B}\}$. The navigation kinematics
is described as follows \cite{hashim2021geometricNAV,hashim2023nonlinear,hashim2023exponentially}:
\begin{equation}
	\left\{ \begin{aligned}\dot{q} & =\frac{1}{2}\Gamma(\omega)q\in\mathbb{S}^{3}\\
		\dot{p} & =v\in\mathbb{R}^{3}\\
		\dot{v} & =g+R_{q}(q)a\in\mathbb{R}^{3}
	\end{aligned}
	\right.\label{c_dyn}
\end{equation}
where
\[
\Gamma(\omega)=\left[\begin{array}{cc}
	0 & -\omega^{\top}\\
	\omega & -[\omega]_{\times}
\end{array}\right]\in\mathbb{R}^{4\times4}
\]
and $g\in\{\mathcal{W}\}$ represents the gravitational acceleration
vector. The Model in \eqref{c_dyn} can be re-formulated as follows:
\begin{equation}
	\left[\begin{array}{c}
		\dot{q}\\
		\dot{p}\\
		\dot{v}\\
		0
	\end{array}\right]=\underbrace{\left[\begin{array}{cccc}
			\frac{1}{2}\Gamma(\omega)q & 0 & 0 & 0\\
			0 & 0 & I_{3} & 0\\
			0 & 0 & 0 & g+R_{q}(q)a\\
			0 & 0 & 0 & 0
		\end{array}\right]}_{M^{c}(q,\omega,a)}\left[\begin{array}{c}
		q\\
		p\\
		v\\
		1
	\end{array}\right]\label{eq:dyn_cont}
\end{equation}
Since the sensor data operates and and is collected in discrete space,
the equation in \eqref{eq:dyn_cont} can be discretized for filter
derivation and implementation. Let the subscript $k$ of variable
$x_{k}$ refers to a sampled signal $x$, $p_{k}\in\mathbb{R}^{3}$
at the $k$th discrete time step. The exact discretized system kinematics
of \eqref{eq:dyn_cont} is given by:
\begin{equation}
	\left[\begin{array}{c}
		q_{k}\\
		p_{k}\\
		v_{k}\\
		1
	\end{array}\right]=\exp(M_{k-1}^{c}dT)\left[\begin{array}{c}
		q_{k-1}\\
		p_{k-1}\\
		v_{k-1}\\
		1
	\end{array}\right]\label{eq:dyn_dis}
\end{equation}
with $M_{k-1}^{c}=M^{c}(q_{k-1},\omega_{k-1},a_{k-1})$ and $dT$
being a sample time.

\subsection{VIN Measurement Model}

\paragraph*{IMU data}

Let $\omega_{m,k}\in\mathbb{R}^{3}$ and $a_{m,k}\in\mathbb{R}^{3}$
denotes the measured angular velocity and linear acceleration at $k$th
time step, respectively, such that
\begin{equation}
	\left\{ \begin{aligned}\omega_{m,k} & =\omega_{k}+b_{\omega,k}+n_{\omega,k}\in\mathbb{R}^{3}\\
		a_{m,k} & =a_{k}+b_{a,k}+n_{a,k}\in\mathbb{R}^{3}\\
		b_{\omega,k} & =b_{\omega,k-1}+n_{b\omega,k-1}\in\mathbb{R}^{3}\\
		b_{a,k} & =b_{a,k-1}+n_{ba,k-1}\in\mathbb{R}^{3}
	\end{aligned}
	\right.\label{a_m}
\end{equation}
where $b_{\omega,k}$ denotes angular velocity bias and $b_{a,k}$
denotes linear acceleration bias. $n_{\omega,k}\in\mathbb{R}^{3}$,
$n_{b\omega,k}$, $n_{a,k}$, and $n_{ba,k}$ describe noise vectors
with zero mean (Gaussian distribution) and $C_{\omega,k}$, $C_{a,k}$,
$C_{b\omega,k}$, and $C_{ba,k}$ covariance matrices. Using \eqref{eq:dyn_dis}
and \eqref{a_m}, define the state vector $x_{k}$:
\begin{equation}
	x_{k}=\begin{bmatrix}q_{k}^{\top} & p_{k}^{\top} & v_{k}^{\top} & b_{\omega,k}^{\top} & b_{a,k}^{\top}\end{bmatrix}^{\top}\in\mathbb{R}^{m_{x}}\label{state}
\end{equation}
with $m_{x}=16$ being the state dimension. Let us introduce the augmented
and additive noise vectors as follows:
\begin{equation}
	\left\{ \begin{aligned}n_{x,k} & =\begin{bmatrix}n_{\omega,k}^{\top} & n_{a,k}^{\top}\end{bmatrix}^{\top}\in\mathbb{R}^{m_{n_{x}}}\\
		n_{w,k} & =\begin{bmatrix}0_{10\times1}^{\top} & n_{b\omega,k}^{\top} & n_{ba,k}^{\top}\end{bmatrix}^{\top}\in\mathbb{R}^{m_{n_{w}}}
	\end{aligned}
	\right.\label{eq:noise vector}
\end{equation}
where $m_{n_{x}}=6$ and $m_{n_{w}}=16$. Let $u_{k}$ be input vector
at time step $k$:
\begin{equation}
	u_{k}=\begin{bmatrix}\omega_{m,k}^{\top} & a_{m,k}^{\top}\end{bmatrix}^{\top}\in\mathbb{R}^{m_{u}}\label{eq:input vector}
\end{equation}
with $m_{u}=6$. Let the augmented state vector be defined as:
\begin{equation}
	x_{k}^{a}=\begin{bmatrix}x_{k}^{\top} & n_{x,k}^{\top}\end{bmatrix}^{\top}\in\mathbb{R}^{m_{a}}\label{eq:augmented state vector}
\end{equation}
with $m_{a}=m_{x}+m_{n_{x}}$. In view of \eqref{eq:dyn_dis}, \eqref{a_m},
\eqref{eq:noise vector}, \eqref{eq:input vector}, and \eqref{eq:augmented state vector},
the overall discrete system kinematics is described by
\begin{equation}
	x_{k}=\operatorname{f}(x_{k-1}^{(i)a},u_{k-1})+n_{w,k-1}\label{state transition}
\end{equation}
with $\operatorname{f}(\cdot):\mathbb{R}^{m_{a}}\times\mathbb{R}^{m_{u}}\rightarrow\mathbb{R}^{m_{x}}$
being the state transition matrix.

\paragraph*{Camera data}

Consider $f_{w,i}\in\mathbb{R}^{3}$ to represent the coordinates
of $i$th landmark point (feature) in $\{\mathcal{W}\}$ extracted
via a series of stereo camera observations at the $k$th time step.
Let $f_{b,i}\in\mathbb{R}^{3}$ denote the $i$th landmark coordinates
in $\{\mathcal{B}\}$ found by triangulating \cite{hartley2003multiple}
the features in the stereo images obtained at time step k. These vectors
are related to each other as follows \cite{hashim2021geometricNAV}:
\begin{align}
	f_{b,i}=R_{q}(q_{k})^{\top}(f_{w,i}-p_{k})+n_{f,i} & \in\mathbb{R}^{3}\label{W2B}
\end{align}
with $n_{f,i}$ being Gaussian white noise related to each landmark
measurement for all $i\in\{1,2,\ldots,m_{f}\}$ and $m_{f}$ being
the total number of landmarks detected at the $k$th sample time.
It is worth noting that landmark points vary among images captured
at different $k$th sample time. Let us define the following relations:
\begin{equation}
	\left\{ \begin{aligned}f_{b} & =\begin{bmatrix}f_{b,1}^{\top} & f_{b,2}^{\top} & \cdots & f_{b,m_{f}}^{\top}\end{bmatrix}^{\top}\in\mathbb{R}^{3m_{f}}\\
		f_{w} & =\begin{bmatrix}f_{w,1}^{\top} & f_{w,2}^{\top} & \cdots & f_{w,m_{f}}^{\top}\end{bmatrix}^{\top}\in\mathbb{R}^{3m_{f}}\\
		n_{f} & =\begin{bmatrix}n_{f,1}^{\top} & n_{f,2}^{\top} & \cdots & n_{f,m_{f}}^{\top}\end{bmatrix}^{\top}\in\mathbb{R}^{3m_{f}}
	\end{aligned}
	\right.\label{eq:feature points}
\end{equation}
From \eqref{eq:feature points}, the expression in \eqref{W2B} can
be reformulated as follows:
\begin{equation}
	f_{b,k}=z_{k}=\operatorname{h}(x_{k},f_{w})+n_{f,k}\in\mathbb{R}^{m_{z}}\label{eq:measurement}
\end{equation}
where
\begin{equation}
	n_{f}\sim N(0,C_{f}\in\mathbb{R}^{m_{z}\times m_{z}})\label{eq:n_f}
\end{equation}
Note that $m_{z}=3m_{f}$.

\section{QUPF-VIN Design\label{sec:Filter}}

\begin{figure}[!h]
	\centering \includegraphics[width=1\columnwidth]{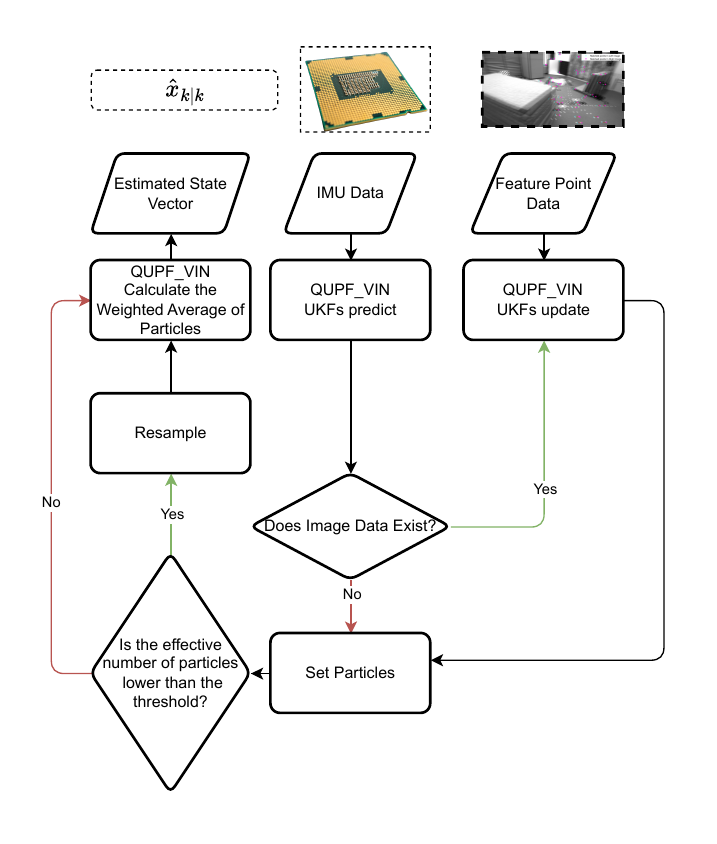}\caption{Illustrative diagram of QUPF-VIN implementation algorithm.}
	\label{fig:flowchart} 
\end{figure}

In this section, the objective is to develop a quaternion-based unscented
particle filter tailored for visual-inertial navigation and applicable
to vehicles travelling in GPS-denied regions. The QUPF-VIN is based
on the UKF \cite{van2000unscented}, modified to operate within the
$\mathbb{S}^{3}$ space and effectively manage the reduced dimensionality
of quaternions. Fig. \ref{fig:flowchart} provides an illustrative
diagram of the proposed QUPF-VIN approach. 

\subsection{QUPF-VIN Initialization}

\subsubsection*{Step 1. Initialization}

QUPF-VIN initialization relies on assigning an initial state estimate
$\hat{x}_{0|0}=\left[\hat{x}_{0|0,q}^{\top},\hat{x}_{0|0,-}^{\top}\right]^{\top}\in\mathbb{R}^{m_{x}}$
and covariance matrix $P_{0|0}$. $\hat{x}_{0|0,q}\in\mathbb{S}^{3}$
and $\hat{x}_{0|0,-}\in\mathbb{R}^{m_{x}-4}$ refer to quaternion
and non-quaternion components, respectively. As straightforward quaternion
subtraction is not feasible, the custom quaternion subtraction presented
in \eqref{eq:q-q} is utilized to enable quaternion subtraction. Consider
the following initialization:
\begin{equation}
	\begin{cases}
		\hat{x}_{0|0} & =\left[\hat{x}_{0|0,q}^{\top},\hat{x}_{0|0,-}^{\top}\right]^{\top}\in\mathbb{R}^{m_{x}}\\
		P_{0|0} & =\text{diag}(P_{\hat{x}_{0|0,q}},P_{0|0,-})\in\mathbb{R}^{(m_{x}-1)\times(m_{x}-1)}
	\end{cases}\label{eq:QNPF_init}
\end{equation}
where $P_{0|0,q}\in\mathbb{R}^{3\times3}$ and $P_{0|0,-}\in\mathbb{R}^{(m_{x}-4)\times(m_{x}-4)}$.

\subsubsection*{Step 2. Particle initialization}

Given the initial covariance and state estimates in \eqref{eq:QNPF_init},
$m_{p}$ particles are conventionally drawn as:
\begin{equation}
	\mathcal{X}_{0}^{(i)}\sim\mathcal{N}(\hat{x}_{0|0},P_{0|0}),\hspace{1em}i=\{1,2,\ldots,m_{p}\}\label{eq:prob conv}
\end{equation}
However, the dimensions of $\hat{x}_{0|0}$, and $P_{0|0}$ are not
compatible (see \eqref{eq:QNPF_init}). Define the following functions
$x^{q2r}(.):\mathbb{R}^{m_{x}}\rightarrow\mathbb{R}^{m_{x}-1}$ and
$x^{r2q}(.):\mathbb{R}^{m_{x}-1}\rightarrow\mathbb{R}^{m_{x}}$ such
that
\begin{align}
	x^{q2r}\left(\begin{bmatrix}x_{q}\\
		x_{-}
	\end{bmatrix}\right) & =\begin{bmatrix}x_{r}\\
		x_{-}
	\end{bmatrix}=\begin{bmatrix}r_{q}(x_{q})\\
		x_{-}
	\end{bmatrix}\in\mathbb{R}^{m_{x}-1}\\
	x^{r2q}\left(\begin{bmatrix}x_{r}\\
		x_{-}
	\end{bmatrix}\right) & =\begin{bmatrix}x_{q}\\
		x_{-}
	\end{bmatrix}=\begin{bmatrix}q_{r}(x_{r})\\
		x_{-}
	\end{bmatrix}\in\mathbb{R}^{m_{x}}
\end{align}
where $x_{q}\in\mathbb{S}^{3}$, $x_{r}\in\mathbb{R}^{3}$, and $x_{-}\in\mathbb{R}^{m_{x}-4}$.
Hence, the expression in \eqref{eq:prob conv} is re-described as
follows:
\begin{equation}
	\left\{ \begin{aligned}\tilde{\mathcal{X}}_{0}^{(i)} & \sim\mathcal{N}\left(x^{q2r}(\hat{x}_{0|0}),P_{0|0}\right)\hspace{0.5cm}i=\{1,2,\ldots,m_{p}\}\\
		\mathcal{X}_{0}^{(i)} & =x^{r2q}(\tilde{\mathcal{X}}_{0}^{(i)})\hspace{2cm}i=\{1,2,\ldots,m_{p}\}
	\end{aligned}
	\right.\label{eq:upf_init}
\end{equation}
where $\tilde{\mathcal{X}}_{0}^{(i)}\in\mathbb{R}^{m_{x}-1}$ and
$\mathcal{X}_{0}^{(i)}\in\mathbb{R}^{m_{x}}$. The weights $w_{0}^{(i)}$
corresponding to each particle are initialized as $\frac{1}{m_{p}}$,
representing equal confidence in all particles. The sigma points mean
$\hat{x}_{0|0}^{(i)}$ and covariance $P_{0|0}^{(i)}$ of each particle
are then initialized as:
\begin{equation}
	\left\{ \begin{aligned}\hat{x}_{0|0}^{(i)} & =\mathcal{X}_{0}^{(i)},\hspace{1em}i=\{1,2,\ldots,m_{p}\}\\
		P_{0|0}^{(i)} & =P_{0|0},\hspace{1em}i=\{1,2,\ldots,m_{p}\}
	\end{aligned}
	\right.\label{eq:ukf_init}
\end{equation}

\subsection{Prediction\label{sec:prediction}}

\subsubsection*{Step 3. Augmentation}

For every $i=\{1,2,\ldots,m_{p}\}$, the mean and covariance estimates
are augmented to capture non-additive noise such that:
\begin{align}
	\hat{x}_{k-1|k-1}^{(i)a} & =\left[\hat{x}_{k-1|k-1}^{(i)^{\top}},0_{m_{n_{x}}\times1}^{\top}\right]^{\top}\in\mathbb{R}^{m_{a}}\label{eq:QNUKF_augment1}\\
	P_{k-1|k-1}^{(i)a} & =\text{diag}(P_{k-1|k-1}^{(i)},C_{x,k})\in\mathbb{R}^{(m_{a}-1)\times(m_{a}-1)}\label{eq:QNUKF_augment2}
\end{align}
where $\hat{x}_{k-1|k-1}^{(i)a}$ and $P_{k-1|k-1}^{(i)a}$ represent
the augmented expected value and covariance matrix of each particle,
respectively. The matrix $C_{x,k}$ representing the covariance matrix
of $n_{x}$ is defined by:
\begin{equation}
	C_{x,k}=\text{diag}(C_{\omega,k},C_{a,k})\in\mathbb{R}^{m_{n_{x}}\times m_{n_{x}}}\label{eq:C_nx}
\end{equation}

\subsubsection*{Step 4. Sigma Point Calculations}

Consider $\delta\hat{x}_{k-1,j}^{(i)a}:=\left(\sqrt{(m_{a}-1+\lambda)P_{k-1|k-1}^{(i)a}}\right)_{j}\in\mathbb{R}^{m_{a}-1}$
with $\lambda\in\mathbb{R}$ being a tuning parameter. It is possible
to divide $\hat{x}_{k-1|k-1}^{(i)a}$ and $\delta\hat{x}_{k-1,j}^{(i)a}$
into their attitude and non-attitude parts $\hat{x}_{k-1|k-1,q}\in\mathbb{S}^{3}$,
$\delta\hat{x}_{k-1,j,r}^{(i)a}\in\mathbb{R}^{3}$, and $\hat{x}_{k-1|k-1,-}^{(i)a}\in\mathbb{R}^{m_{a}-4}$,
$\delta\hat{x}_{k-1,j,-}^{(i)a}\in\mathbb{R}^{m_{a}-4}$, as outlined
below:
\begin{equation}
	\left\{ \begin{aligned}\hat{x}_{k-1|k-1}^{(i)a} & =\left[(\hat{x}_{k-1|k-1,q}^{(i)a})^{\top},(\hat{x}_{k-1|k-1,-}^{(i)a})^{\top}\right]^{\top}\\
		\delta\hat{x}_{k-1,j}^{(i)a} & =\left[(\delta\hat{x}_{k-1,j,r}^{(i)a})^{\top},(\delta\hat{x}_{k-1,j,-}^{(i)a})^{\top}\right]^{\top}
	\end{aligned}
	\right.\label{eq:693}
\end{equation}
In the light of \eqref{eq:q+r}, \eqref{eq:q-r}, and \eqref{eq:693},
the sigma points of the $i$th UKF of QUPF-VIN are found by:
\begin{equation}
	\left\{ \begin{aligned}\mathcal{X}_{k-1|k-1,0}^{(i)a} & =\hat{x}_{k-1|k-1}^{(i)a}\in\mathbb{R}^{m_{a}}\\
		\mathcal{X}_{k-1|k-1,j}^{(i)a} & =\hat{x}_{k-1|k-1}^{(i)a}\oplus\delta\hat{x}_{j}^{(i)a}\\
		& =\begin{bmatrix}\hat{x}_{k-1|k-1,q}^{(i)a}\oplus\delta\hat{x}_{k-1,j,r}^{(i)a}\\
			\hat{x}_{k-1|k-1,-}^{(i)a}+\delta\hat{x}_{k-1,j,-}^{(i)a}
		\end{bmatrix}\in\mathbb{R}^{m_{a}}\\
		\mathcal{X}_{k-1|k-1,j+m_{a}}^{(i)a} & =\hat{x}_{k-1|k-1}^{(i)a}\ominus\delta\hat{x}_{j}^{(i)a}\\
		& =\begin{bmatrix}\hat{x}_{k-1|k-1,q}^{(i)a}\ominus\delta\hat{x}_{k-1,j,r}^{(i)a}\\
			\hat{x}_{k-1|k-1,-}^{(i)a}-\delta\hat{x}_{k-1,j,-}^{(i)a}
		\end{bmatrix}\in\mathbb{R}^{m_{a}},\\
		& \hspace{1.5cm}j=\{1,2,\ldots,2(m_{a}-1)\}
	\end{aligned}
	\right.\label{eq:Sigma_QNUKF}
\end{equation}

\subsubsection*{Step 5. Propagation}

Given IMU measurements, each sigma point for each UKF is propagated
through the state transition function \eqref{state transition} to
find predicted sigma points $\mathcal{X}_{k|k-1,j}^{(i)}$. This can
be shown as:
\begin{equation}
	\mathcal{X}_{k|k-1,j}^{(i)}=\operatorname{f}(\mathcal{X}_{k-1|k-1,j}^{(i)a},u_{k-1})\in\mathbb{R}^{m_{x}}\label{eq:propagate_sigma_points_augmented}
\end{equation}
Using the propagated sigma points, the mean $\hat{x}_{k|k-1}^{(i)}$
and covariance matrix $P_{k|k-1}^{(i)}$ for each UKF should be computed.
Consider:
\begin{align}
	\hat{x}_{k|k-1}^{(i)} & =\begin{bmatrix}\hat{x}_{k|k-1,q}^{(i)^{\top}} & \hat{x}_{k|k-1,-}^{(i)^{\top}}\end{bmatrix}^{\top}\in\mathbb{R}^{m_{x}}\label{eq:x_k_k_1}\\
	\mathcal{X}_{k|k-1,j}^{(i)} & =\begin{bmatrix}\mathcal{X}_{k|k-1,q}^{(i)^{\top}} & \mathcal{X}_{k|k-1,-}^{(i)^{\top}}\end{bmatrix}^{\top}\in\mathbb{R}^{m_{x}}\label{eq:QNUKF_Xkk-1}
\end{align}
where $\hat{x}_{k|k-1,q}^{(i)},\mathcal{X}_{k|k-1,q}^{(i)}\in\mathbb{S}^{3}$,
and $\hat{x}_{k|k-1,-}^{(i)},\mathcal{X}_{k|k-1,-}^{(i)}\in\mathbb{R}^{m_{x}-4}$.
Thereby, one has
\begin{align}
	\hat{x}_{k|k-1}^{(i)} & =\begin{bmatrix}\operatorname{QWA}(\{\mathcal{X}_{k|k-1,j,q}^{(i)}\},\{w_{j}^{m}\})\\
		{\displaystyle \sum_{j=0}^{2(m_{a}-1)}w_{j}^{m}\mathcal{X}_{k|k-1,j,-}^{(i)}}
	\end{bmatrix}\in\mathbb{R}^{m_{x}}\label{eq:state mean - QNUKF}
\end{align}
{\small{}
	\begin{align}
		P_{k|k-1}^{(i)} & =\sum_{j=0}^{2(m_{a}-1)}w_{j}^{c}\left(\mathcal{X}_{k|k-1,j}^{(i)}\ominus\hat{x}_{k|k-1}^{(i)}\right)\left(\mathcal{X}_{k|k-1,j}^{(i)}\ominus\hat{x}_{k|k-1}^{(i)}\right)^{\top}\nonumber \\
		& \qquad+C_{w,k}\in\mathbb{R}^{(m_{x}-1)\times(m_{x}-1)}\label{eq:state cov - QNUKF}
	\end{align}
}where
\begin{equation}
	C_{w,k}=\begin{bmatrix}0_{9\times9} & 0_{3\times3} & 0_{3\times3}\\
		0_{9\times9} & C_{b\omega,k} & 0_{3\times3}\\
		0_{9\times9} & 0_{3\times3} & C_{ba,k}
	\end{bmatrix}\in\mathbb{R}^{(m_{n_{w}}-1)\times(m_{n_{w}}-1)}\label{eq:C_nw}
\end{equation}
Considering \eqref{eq:q-q}, the subtraction in \eqref{eq:state cov - QNUKF}
is obtained by:
\begin{align}
	\mathcal{X}_{k|k-1,j}^{(i)}\ominus\hat{x}_{k|k-1}^{(i)} & =\begin{bmatrix}\mathcal{X}_{k|k-1,j,q}^{(i)}\ominus\hat{x}_{k|k-1,q}^{(i)}\\
		\mathcal{X}_{k|k-1,j,-}^{(i)}-\hat{x}_{k|k-1,-}^{(i)}
	\end{bmatrix}\in\mathbb{R}^{m_{x}-1}\label{eq:x-x}
\end{align}
Note that the weights $w_{j}^{m}$ and $w_{j}^{c}$ in \eqref{eq:state mean - QNUKF}
and \eqref{eq:state cov - QNUKF} are found by:
\begin{equation}
	\left\{ \begin{aligned}w_{0}^{m} & =\frac{\lambda}{\lambda+(m_{a}-1)}\in\mathbb{R}\\
		w_{0}^{c} & =\frac{\lambda}{\lambda+(m_{a}-1)}+1-\alpha^{2}+\beta\in\mathbb{R}\\
		w_{j}^{m} & =w_{j}^{c}=\frac{1}{2((m_{a}-1)+\lambda)}\in\mathbb{R}\\
		& \hspace{2cm}j=\{1,\ldots,2(m_{a}-1)\}
	\end{aligned}
	\right.\label{eq:weights-QNUKF}
\end{equation}
with $\alpha,\beta\in\mathbb{R}$ being tuning parameters.

\subsection{Update\label{sec:update}}

\subsubsection*{Step 6. Predict Measurement}

Every sigma point is passed through the measurement function \eqref{eq:measurement}
to predict the measurement vector. The measurement sigma points $\mathcal{Z}_{k|k-1,j}^{(i)}$
are obtained as follows:
\begin{align}
	\mathcal{Z}_{k|k-1,j}^{(i)} & =h(\mathcal{X}_{k|k-1,j}^{(i)},f_{w})\in\mathbb{R}^{m_{z}}\label{eq:propagate measurment}
\end{align}
The covariance matrices $P_{z_{k},z_{k}}^{(i)}\in\mathbb{R}^{m_{z}\times m_{z}}$
and $P_{x_{k},z_{k}}^{(i)}\in\mathbb{R}^{(m_{x}-1)\times m_{z}}$
and the mean estimated measurement vector $\hat{z}_{k|k-1}^{(i)}\in\mathbb{R}^{m_{z}}$
for each UKF are found by:{\small{}
	\begin{align}
		\hat{z}_{k|k-1}^{(i)} & =\sum_{j=0}^{2(m_{a}-1)}w_{j}^{m}\mathcal{Z}_{k|k-1,j}^{(i)}\label{eq:zhat - QNUKF}\\
		P_{z_{k},z_{k}}^{(i)} & =\sum_{j=0}^{2(m_{a}-1)}w_{j}^{c}[\mathcal{Z}_{k|k-1,j}^{(i)}-\hat{z}_{k|k-1}^{(i)}][\mathcal{Z}_{k|k-1,j}^{(i)}-\hat{z}_{k|k-1}^{(i)}]^{\top}\nonumber \\
		& \hspace{1.5cm}+C_{f}\label{eq:P_zz - QUKF}\\
		P_{x_{k},z_{k}}^{(i)} & =\sum_{j=0}^{2(m_{a}-1)}w_{j}^{c}[\mathcal{X}_{k|k-1,j}^{(i)}\ominus\hat{x}_{k|k-1}^{(i)}][\mathcal{Z}_{k|k-1,j}^{(i)}-\hat{z}_{k|k-1}^{(i)}]^{\top}\label{eq:P_xz - QUKF}
	\end{align}
}The $\ominus$ operator in \eqref{eq:P_xz - QUKF} follows the map
in \eqref{eq:x-x}. The Kalman gains $K_{k}^{(i)}$, estimation covariance
matrices $P_{k|k}^{(i)}$ and correction vectors $\delta\hat{x}_{k|k-1}^{(i)}$
are defined by:
\begin{align}
	K_{k}^{(i)} & =P_{x_{k},z_{k}}^{(i)}P_{z_{k},z_{k}}^{(i)^{\top}}\in\mathbb{R}^{(m_{x}-1)\times m_{z}}\label{eq:K}\\
	P_{k|k}^{(i)} & =P_{k|k-1}^{(i)}-K_{k}^{(i)}P_{z_{k},z_{k}}^{(i)}K_{k}^{(i)^{\top}}\in\mathbb{R}^{(m_{x}-1)\times(m_{x}-1)}\label{eq:update:P}\\
	\delta\hat{x}_{k|k-1}^{(i)}: & =K_{k}^{(i)}(z_{k}-\hat{z}_{k|k-1}^{(i)})\in\mathbb{R}^{m_{x}-1}\label{eq:correction vector}
\end{align}
Let us divide $\delta\hat{x}_{k|k-1}^{(i)}$ into its attitude ($\delta\hat{x}_{k|k-1,r}^{(i)}\in\mathbb{R}^{3}$)
and non-attitude ($\delta\hat{x}_{k|k-1,-}^{(i)}\in\mathbb{R}^{m_{x}-4}$)
components:
\begin{equation}
	\delta\hat{x}_{k|k-1}^{(i)}=\begin{bmatrix}\delta\hat{x}_{k|k-1,r}^{(i)^{\top}} & \delta\hat{x}_{k|k-1,-}^{(i)^{\top}}\end{bmatrix}^{\top}\label{eq:delx_k_k_1}
\end{equation}
Then, the estimated state vector for each UKF $\hat{x}_{k|k}^{(i)}$
is defined by:
\begin{equation}
	\hat{x}_{k|k}^{(i)}=\hat{x}_{k|k-1}^{(i)}\oplus\delta\hat{x}_{k|k-1}^{(i)}\label{eq:final update - QUNKF}
\end{equation}
where
\begin{equation}
	\hat{x}_{k|k-1}^{(i)}\oplus\delta\hat{x}_{k|k-1}^{(i)}=\begin{bmatrix}\hat{x}_{k|k-1,q}^{(i)}\oplus\delta\hat{x}_{k|k-1,r}^{(i)}\\
		\hat{x}_{k|k-1,-}^{(i)}+\delta\hat{x}_{k|k-1,-}^{(i)}
	\end{bmatrix}\label{eq:QNUKF_correction}
\end{equation}

\subsubsection*{Step 7. Particle and Weight calculations}

Using the estimated vector $\hat{x}_{k|k-1}^{(i)}$ and covariance
matrix $P_{k|k}^{(i)}$ as the mean and covariance matrix of a Gaussian
distribution, the particles $\mathcal{X}_{k}^{(i)}$ are drawn similar
to \eqref{eq:upf_init}:
\begin{equation}
	\left\{ \begin{aligned}\tilde{\mathcal{X}}_{k}^{(i)} & \sim\mathcal{N}\left(x^{q2r}\left(\hat{x}_{k|k-1}^{(i)}\right),P_{k|k}^{(i)}\right)\\
		\mathcal{X}_{k}^{(i)} & =x^{r2q}(\tilde{\mathcal{X}}_{k}^{(i)})
	\end{aligned}
	\right.\label{eq:upf_particles}
\end{equation}
The weights $w_{k}^{(i)}$ corresponding to each particle at the current
time-step represent how accurate each particle is and they are defined
by:
\begin{equation}
	\tilde{w}_{k}^{(i)}=\frac{\mathbb{P}\left(z_{k}|\mathcal{X}_{k}^{(i)}\right)\mathbb{P}\left(\mathcal{X}_{k}^{(i)}|\mathcal{X}_{k-1}^{(i)}\right)}{\mathbb{P}\left(\mathcal{X}_{k}^{(i)}|z_{1:k}\right)+\epsilon}+\epsilon\label{eq:weight_tild}
\end{equation}
$\epsilon$ is added to avoid numerical instabilities for the case
of zero weight and division by zero. The probability terms in \eqref{eq:weight_tild}
are calculated by:{\small{}
	\begin{equation}
		\begin{cases}
			\mathbb{P}\left(z_{k}|\mathcal{X}_{k}^{(i)}\right) & =\mathcal{N}\left(z_{k}|\operatorname{h}\left(\mathcal{X}_{k}^{(i)}\right),C_{f}\right)\\
			\mathbb{P}\left(\mathcal{X}_{k}^{(i)}|\mathcal{X}_{k-1}^{(i)}\right) & =\mathcal{N}\left(x^{q2r}\left(\mathcal{X}_{k}^{(i)}\right)|x^{q2r}\left(\hat{x}_{k|k-1}^{(i)}\right),P_{k|k-1}^{(i)}\right)\\
			\mathbb{P}\left(\mathcal{X}_{k}^{(i)}|z_{1:k}\right) & =\mathcal{N}\left(x^{q2r}\left(\mathcal{X}_{k}^{(i)}\right)|x^{q2r}\left(\hat{x}_{k|k}^{(i)}\right),P_{k|k}^{(i)}\right)
		\end{cases}\label{eq:combined_probabilities}
	\end{equation}
}Next, the weights are normalized as follows:
\begin{equation}
	w_{k}^{(i)}=\frac{\tilde{w}_{k}^{(i)}}{{\displaystyle \sum_{i=1}^{m_{p}}\tilde{w}_{k}^{(i)}}}
\end{equation}

\subsubsection*{Step 8. Resampling\label{sec:resampling}}

To address the degeneracy challenging problem of particle filters,
resampling is performed once the effective number of samples $m_{eff}\in\mathbb{R}$
falls below a predefined threshold $m_{thr}\in\mathbb{R}$ \cite{fu2010improvement}.
The effective number of samples is calculated by:
\begin{equation}
	m_{eff}=\frac{1}{\sum\left(w_{k}^{(i)}\right)^{2}}
\end{equation}
The particles are then resampled if $m_{eff}$ is lower than a certain
threshold $m_{thr}$. Consider the set $\left\{ \mathcal{X}_{k}^{(i)},P_{k|k}^{(i)}\right\} $
as instances of a random variable, associated with the probability
set $\{w_{k}^{(i)}\}$. During the resampling step, $m_{p}$ samples
are drawn from $\left\{ \mathcal{X}_{k}^{(i)},P_{k|k}^{(i)}\right\} $
according to their corresponding probabilities. Note that a single
particle may be sampled multiple times. After resampling, the weights
are updated to reflect a uniform distribution, as the distribution
is now represented by the number of particles rather than their individual
weights. The resampling process is formally expressed as:{\small{}
	\begin{equation}
		\left[\left\{ \mathcal{X}_{k}^{(i)},P_{k|k}^{(i)}\right\} ,\{w_{k}^{(i)}\}\right]\leftarrow\operatorname{Resample}\left(\left\{ \mathcal{X}_{k}^{(i)},P_{k|k}^{(i)}\right\} ,\{w_{k}^{(i)}\right\} )\label{eq:resample}
	\end{equation}
}{\small\par}

\subsubsection*{Step 9. Particles weighted average}

The weighted average of the particles will be the estimated state
vector at the current time step. Let us divide each particle into
its quaternion $\mathcal{X}_{k,q}\in\mathbb{S}^{3}$ and non-quaternion
$\mathcal{X}_{k,-}$ components. Hence, the estimated state vector
$\hat{x}_{k|k}$ is defined by:
\begin{equation}
	\hat{x}_{k|k}=\begin{bmatrix}\operatorname{QWA}\left(\left\{ \mathcal{X}_{k,q}^{(i)}\right\} ,\left\{ w_{k}^{(i)}\right\} \right)\\
		{\displaystyle \sum_{i=0}^{m_{p}}w_{k}^{(i)}\mathcal{X}_{k,-}^{(i)}}
	\end{bmatrix}\in\mathbb{R}^{m_{x}}
\end{equation}
The particles will also be set as the expected value of the UKFs’
estimated state vectors which will be used in the next iteration at
\ref{eq:QNUKF_augment1}. In other words:
\[
x_{k|k}^{(i)}=\mathcal{X}_{k}^{(i)}
\]

\subsubsection*{Step 10. Iterate}

Go back to Step 2 and iterate with $k\rightarrow k+1$.

\section{Numerical Results\label{sec:Results}}

\begin{figure}[h]
	\centering \includegraphics[width=1\columnwidth]{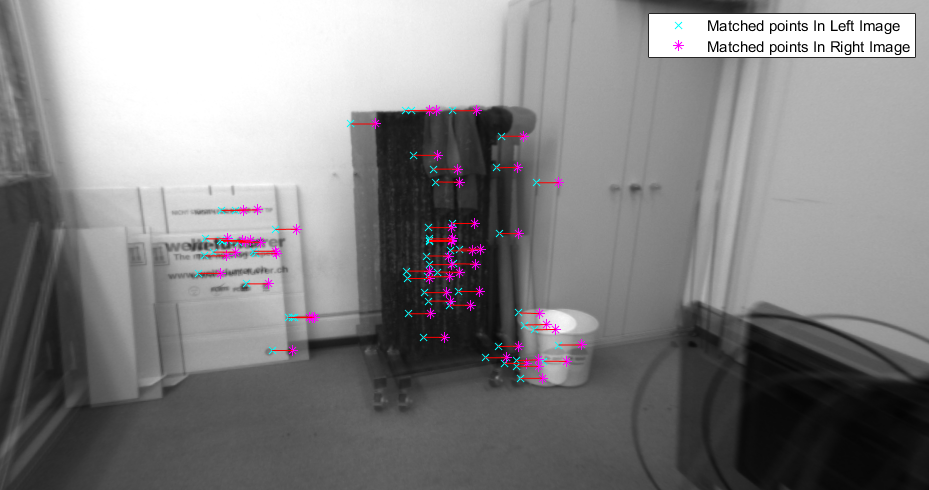}
	\caption{A sample of matched landmark data points from left to right frame
		of EuRoC dataset \cite{Burri25012016}.}
	\label{fig:matched} 
\end{figure}

This section evaluates the effectiveness and robustness of the proposed
QUPF-VIN algorithm using a real-world dataset from a quadrotor flight
in 3D space, specifically the EuRoC dataset \cite{Burri25012016}.
The test platform is the Asctec Firefly hex-rotor Micro Aerial Vehicle
(MAV), operating in a GPS-denied indoor environment. Ground truth
data, including true position and orientation (quaternion), were collected
using an OptiTrack localization system. The measurements consist of
6-axis IMU data (linear acceleration and angular velocity) and stereo
images. The stereo images, captured at 20 Hz, were obtained from an
Aptina MT9V034 global shutter sensor, while the IMU data, including
linear acceleration and angular velocity, were collected at 200 Hz
using an ADIS16448 sensor. Due to the difference in sampling rates
between the IMU and the camera, landmark measurements are not available
for every IMU data point. To address this challenge, the proposed
algorithm updates the state when image data is available. Otherwise,
the particles $\mathcal{X}_{k}^{(i)}$ are set to the predicted state
vector $\hat{x}_{k|k-1}^{(i)}$ while image data is unavailable.

\begin{figure*}[!t]
	\centering \includegraphics[width=2\columnwidth]{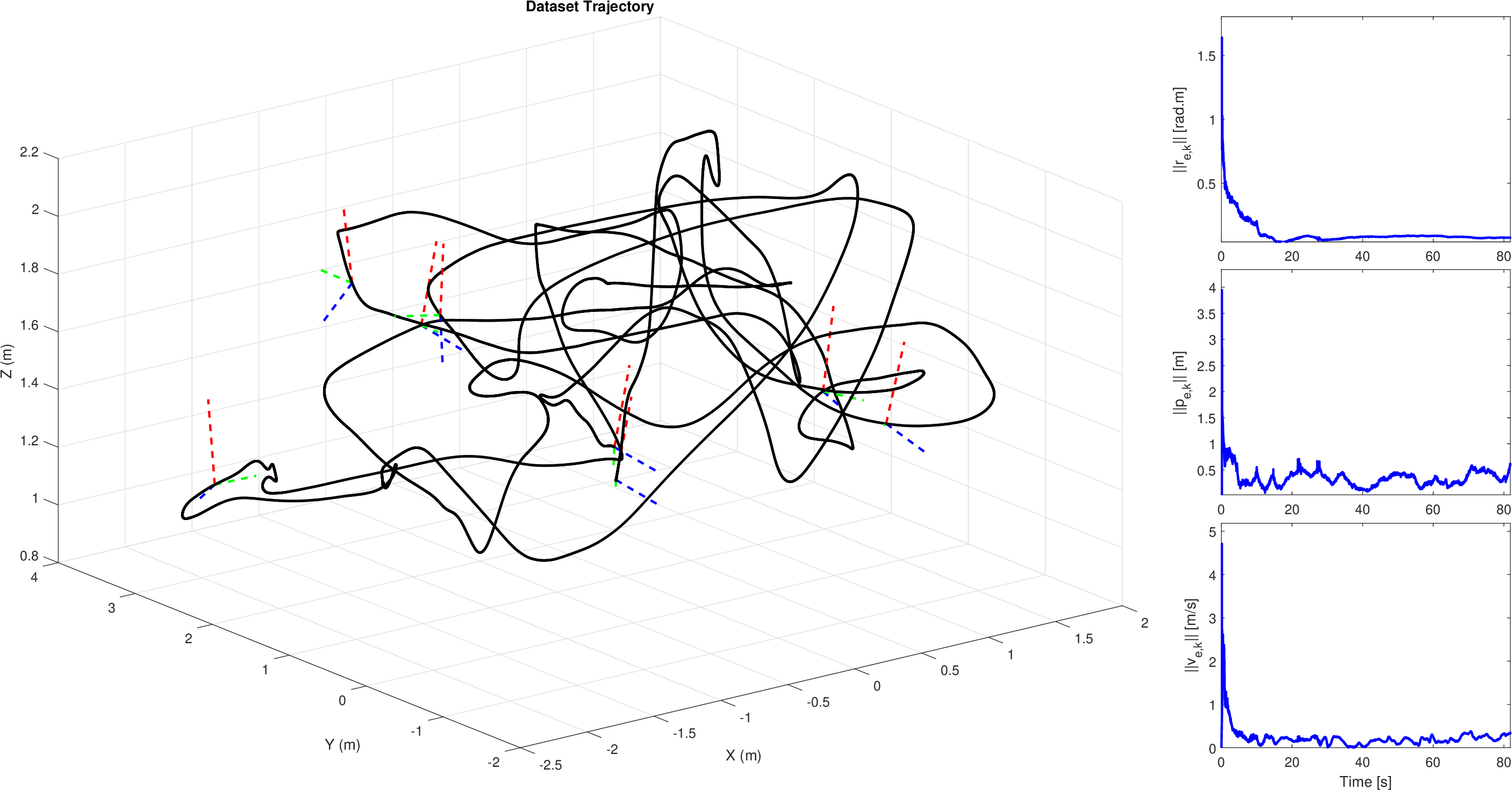} \caption{Performance assessment using the EuRoC V1\_02\_medium dataset \cite{Burri25012016}.
		The left side shows UAV navigation (estimation) trajectory 3D space
		where the position is depicted in black solid line while the orientation
		is represented by red, green, and blue dashed lines. The right side
		presents normalized values of error vectors: orientation error $\|r_{e,k}\|$,
		position error $\|p_{e,k}\|$, and linear velocity error $\|v_{e,k}\|$
		in blue solid lines.}
	\label{fig:summary} 
\end{figure*}

\begin{figure}[!h]
	\centering \includegraphics[width=1\columnwidth]{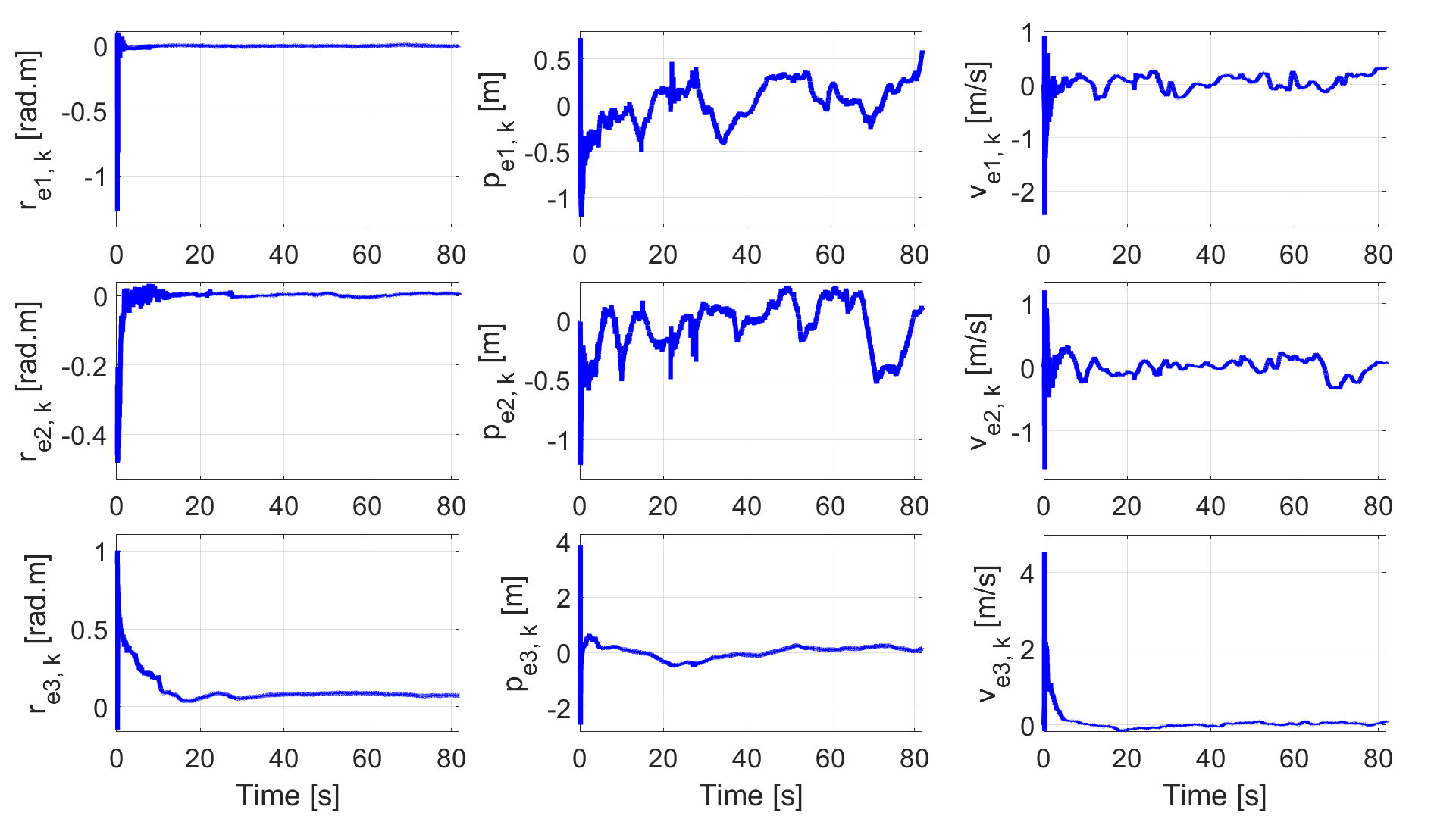}
	\caption{Estimation error: Rotation (left portion), position (middle portion),
		and linear velocity (rightp protion).}
	\label{fig:errors} 
\end{figure}

\begin{figure}[!h]
	\centering \includegraphics[width=1\columnwidth]{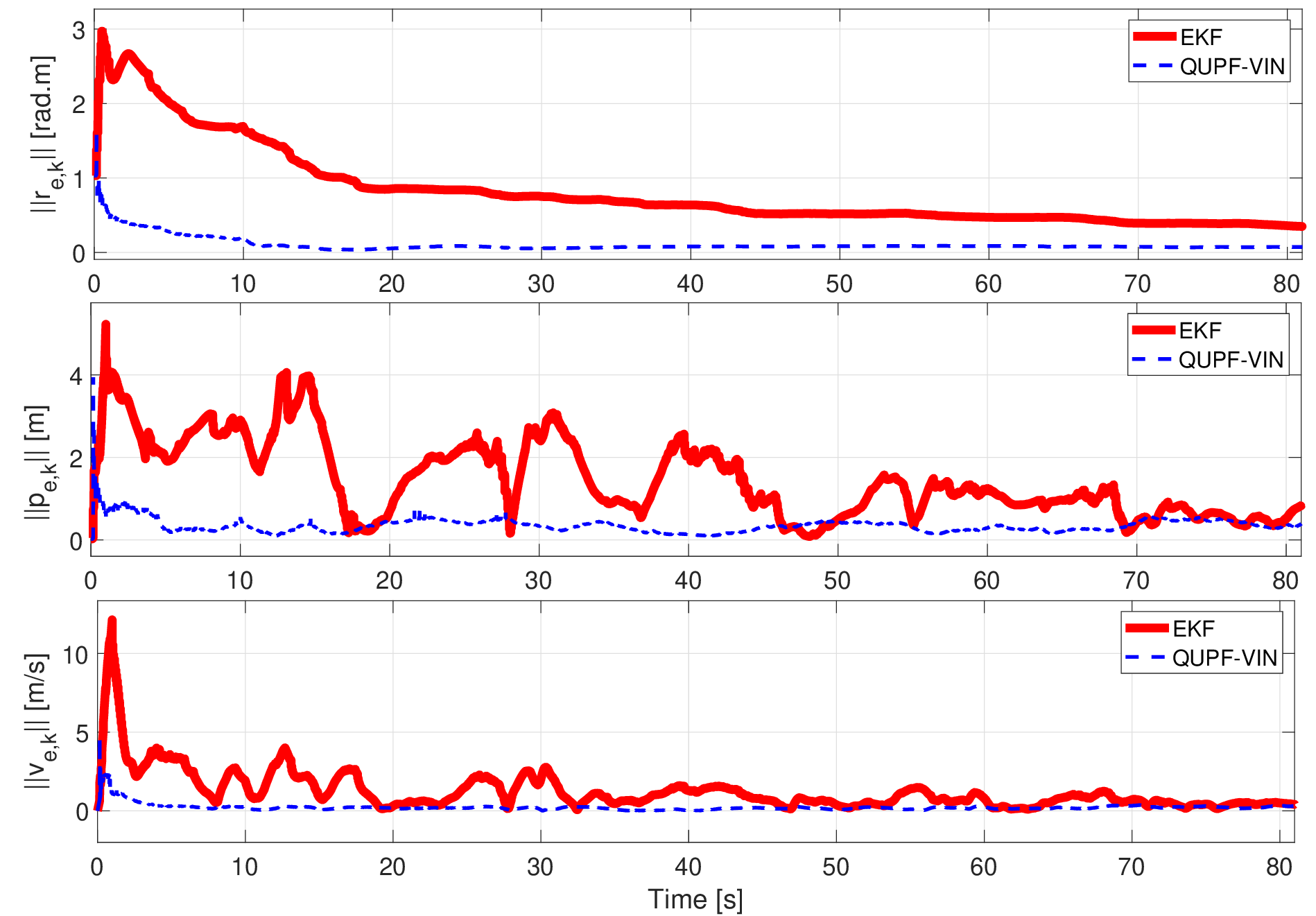} \caption{Comparison results of EKF (literature in red) and the proposed QUPF-VIN
		(in blue).}
	\label{fig:ekf_upf} 
\end{figure}

For every set of stereo images, the landmark points are defined via
the Kanade-Lucas-Tomasi (KLT) approach \cite{shi1994good}. As illustrative
example, the landmark matching between two instantaneous frame is
presented in Fig. \ref{fig:matched}. The mapping triangulation approach
in \cite{hartley2003multiple} were utilized to project the 2D matched
points into the 3D space, describing the landmark pointThe filter
was also compared to the EKF, which is a commonly adopted base filter
in this domain. To ensure a fair comparison, both filters were initialized
with the same values and parameters. In Fig. \ref{fig:ekf_upf} the
magnitude of orientation (top), position (middle), and velocity (bottom)
estimation errors are plotted against time. The EKF results are represented
by solid red lines, while the QUPF-VIN results are depicted by dashed
blue lines. As shown in Fig. \ref{fig:ekf_upf}, the proposed filter
outperformed the EKF in terms of accuracy and speed, specifically
in reducing the magnitudes of orientation, position, and linear velocity
estimation errors.s in $\{\mathcal{W}\}$. The mapping $\mathbb{S}^{3}\times\mathbb{S}^{3}\rightarrow\mathbb{R}^{3}$
associated with the subtraction operator provided in \eqref{eq:q-q}
is used to define the orientation estimation error $r_{e,k}$ such
that: 
\begin{equation}
	r_{e,k}=q_{k}\ominus\hat{q}_{k}\label{eq:r_e}
\end{equation}
with $r_{e,k}=\begin{bmatrix}r_{e1,k} & r_{e2,k} & r_{e3,k}\end{bmatrix}^{\top}\in\mathbb{R}^{3}$.
Consider expressing the estimation errors of position and linear velocity
at the $k$th sample step as follows: 
\begin{align}
	p_{e,k} & =p_{k}-\hat{p}_{k}=\begin{bmatrix}p_{e1,k} & p_{e2,k} & p_{e3,k}\end{bmatrix}^{\top}\in\mathbb{R}^{3}\label{eq:p_e}\\
	v_{e,k} & =v_{k}-\hat{v}_{k}=\begin{bmatrix}v_{e1,k} & v_{e2,k} & v_{e3,k}\end{bmatrix}^{\top}\in\mathbb{R}^{3}\label{eq:v_e}
\end{align}
Fig. \ref{fig:summary} presents the performance of QUPF-VIN using
the EuRoC V1\_02\_medium room dataset \cite{Burri25012016}. The left
portion of Fig. \ref{fig:summary} shows the drone's estimated position
trajectory and orientation during the navigation experiment with 6
DoF. The right portion of Fig. \ref{fig:summary} reveals the estimation
errors for orientation, position, and linear velocity. As illustrated
in Fig. \ref{fig:summary}, the proposed algorithm exhibits rapid
error convergence to near-zero values, even when initialized with
large errors, confirming the robustness and reliability of the QUPF-VIN
algorithm. This confirms the robustness and reliability of the proposed
QUPF-VIN algorithm. To further evaluate the filter's performance,
Fig. \ref{fig:errors} plots each component of the orientation, position,
and linear velocity estimation errors over time, demonstrating consistent
convergence across all dimensions.

Additionally, the filter was compared to the EKF, a widely used baseline
in this field. For a fair comparison, both filters were initialized
with identical values and parameters. In Fig. \ref{fig:ekf_upf},
the magnitudes of the orientation (top), position (middle), and velocity
(bottom) estimation errors are plotted against time, with EKF results
represented by solid red lines and QUPF-VIN results by dashed blue
lines. As shown in Fig. \ref{fig:ekf_upf}, the proposed filter outperforms
the EKF in both accuracy and speed, significantly reducing the magnitude
of the orientation, position, and linear velocity estimation errors.

\section{Conclusion\label{sec:Conclusion}}

This article investigated the navigation problem of a vehicle operating
with six degrees of freedom. A novel geometric Quaternion-based Unscented
Particle Filter for Visual-Inertial Navigation (QUPF-VIN) has been
developed to estimate the vehicle's navigation state (orientation,
position, and linear velocity) while mitigating measurement uncertainties.
The proposed filter effectively addressed kinematic nonlinearities
and ensures computational efficiency, even at low sampling rates.
The proposed algorithm has been structured using unit quaternions
to accurately model true navigation kinematics and avoid singularities.
The algorithm leveraged sensor fusion from a vision unit (e.g., monocular
or stereo camera) and a 6-axis IMU. The performance of the QUPF-VIN
was evaluated using a real-world dataset of an indoor drone flight,
which included stereo camera images and IMU data collected at a low
sampling rate. The results demonstrated good navigation performance,
with tracking errors approaching zero. Furthermore, the proposed filter
outperformed a baseline EKF in comparison.

\bibliographystyle{IEEEtran}
\bibliography{bib_QUKF_ACC}

\end{document}